\documentstyle[aclap,german,psfig]{article}

\newfont{\tablefont}{cmtt9}

\title{An Approach to Proper Name Tagging for German}
\author{Christine Thielen\\Seminar f\"ur Sprachwissenschaft\\
	Universit\"at T\"ubingen\\Wilhelmstr. 113\\D--72074 T\"ubingen\\
	Germany\\e-mail: thielen@sfs.nphil.uni-tuebingen.de}
\date{\today}

\begin{document}

\maketitle

\begin{abstract}

This paper presents an incremental method for the tagging of proper
names in German newspaper texts. The tagging is performed by the
analysis of the syntactic and textual contexts of proper names
together with a morphological analysis. The proper names selected by
this process supply new contexts which can be used for finding new
proper names, and so on. This procedure was applied to a small German
corpus (50,000 words) and correctly disambiguated 65\% of the
capitalized words, which should improve when it is applied to a very
large corpus.

\end{abstract}

\section{Introduction}

The recognition of proper names constitutes one of the major problems
for the wealth of tagging systems developed in the last few
years. Most of these systems are statistically based and make use of
statistical properties which are acquired from a large manually tagged
training corpus. The formation of new proper names, especially
personal names, is very productive, and it is not feasible to list
them in a static lexicon. As Church \cite{Church1988} already
discussed for English, it is difficult to decide whether a capitalized
word is a proper name if it has a low frequency (\( < \) 20), and so
they were removed from the lexicon. But because they are highly
individual, this is the case for most proper names. Furthermore, the
problem of proper name tagging for German is not restricted to the
disambiguation of sentence--initial words, because proper names and
generic terms (normal nouns) are capitalized both at the beginning and
within a sentence. Church suggested labelling words as proper nouns if
they are ``adjacent to'' other capitalized words. This also holds for
German proper nouns, but it is difficult to decide which of the
capitalized words belong to the proper name and which not, e.g. is it
a first name (as in ``Helmut Kohl'') or is it an apposition (as in
``Bundeskanzler Kohl''), or is it a complex institutional name
composed of several generic terms and a proper name (as in
``Vereinigte Staaten von Amerika''). In this procedure, I use Church's
heuristic for the selection of proper name hypotheses, which are
evaluated on the basis of their syntactic and textual context together
with a morphological analysis.  The starting point of the analysis is
a small database of definite minimal contexts like titles
(e.g. ``Prof.'', ``Dr.'') and forms of address (e.g. ``Herr'',
``Frau''), which increases with the processing of texts in which
proper names are identified, and supplies new contexts which can be
used to find new proper names and new contexts, etc.. This
incremental method is applied to unrestricted texts of a small corpus
(50,000 words) of German newspapers.

\section{Proper Name Acquisition}

{}From a psycholinguistic point of view it is possible that we memorize
proper names better if we organize them in a hierarchy, in which each word
would constitute a node whose subordinate nodes are its hyponyms
\cite{Koss1990}. For example, we find in the semantic hierarchy in
figure 1 SOCRATES as hyponym of PHILOSOPHER and PHILOSOPHER as hyponym
of SCHOLAR, and each node may bear features describing properties of
the node.

\begin{figure}[htbf]
\centerline{\psfig{figure=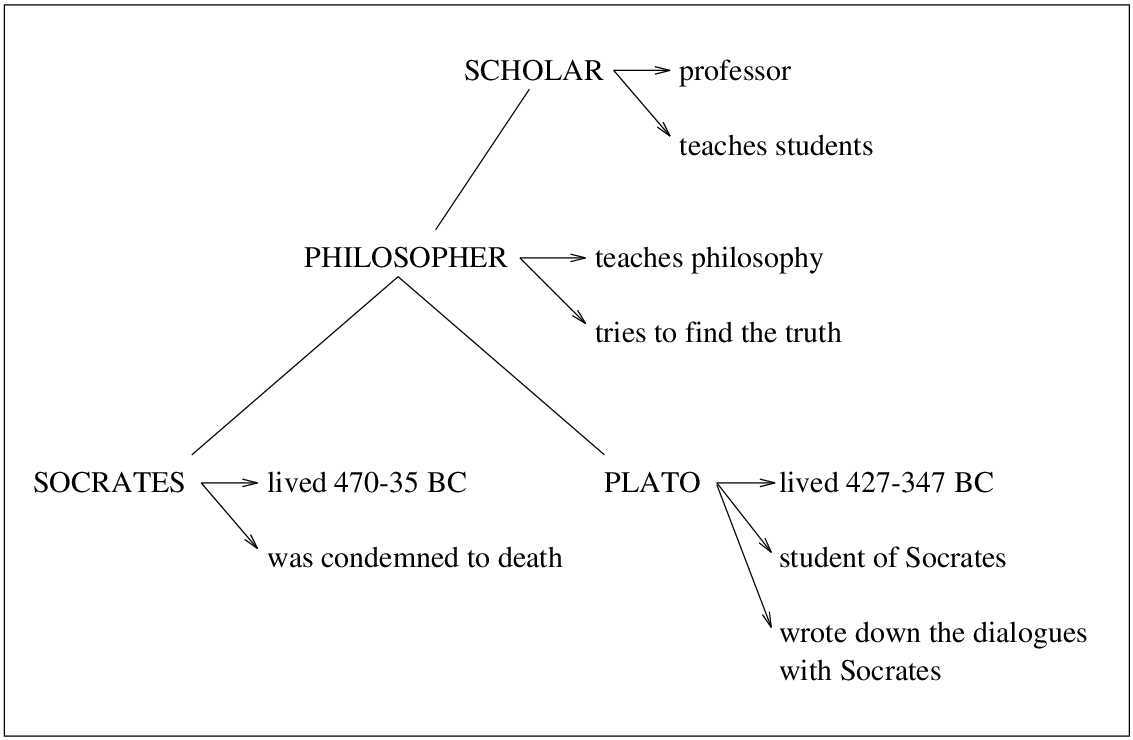,width=7.5cm}}
\captionsenglish{{\small Figure 1: SOCRATES in a semantic hierarchy}}
\label{fig:semnet}
\end{figure}

One can observe that hyperonyms of names are used to identify or to
introduce a proper name in texts. If the knowledge of a name cannot be
presupposed, then the name is often introduced by an appositional
construction (1)-(2) \cite{Hackel1986} and can be used without
additional information (3)--(4) \cite{Kalverkamper1978} later on.

\begin{itemize}
\item[(1)] der Vorsitzende des Verteidigungsausschusses, {\em Biehle}
		(CSU), hat Verteidigungsminister {\em W\"or\-ner} gebeten, ...\\
	(the chair of the defence committee, {\em Biehle} (CSU), asked the Minister of
Defence {\em W\"orner} to ...)
\item[(2)] der SPD--Abgeordnete {\em Gerster} kritisierte, da{\ss} ...\\
	(the SPD member of parliament {\em Gerster} criticized that ...)
\item[(3)] In einem Fernschreiben an {\em W\"orner}, \"au{\ss}erte {\em Biehle}
		 am Dienstag, ...\\
	(in a telex to {\em W\"orner}, {\em Biehle} commented on Tuesday ...)
\item[(4)] {\em Gerster} forderte eine Mindestflugh\"ohe von 300 Metern\\
	({\em Gerster} called for a minimal flying height of 300 metres)
\end{itemize}

The syntactic analysis (see section \ref{sec:entag}) operates on a
small lexicon of definite minimal contexts of proper names
(MC--lexicon) which are used in such appositional constructions and
generates a lexicon of so--called potential minimal contexts
(MCpot--lexicon).

In addition there exist other methods \cite{Koss1987} for the
acquisiton of proper names, two of which can be directly observed in
the texts. The first method (``Lernpsychologische Sinnverleihung'')
tries to lend sense to the name in order to learn it, e.g. the name
``D\"usseldorf'' is given the meaning of `village'. Today it is a big
city, but the compound part {\em -dorf} helps us to identify it as a
proper name. The second method, the formation of name fields
(``Namenfelder'') and name scenes (``Namenlandschaften''), helps us to
recognize names describing places which belong to a certain district
or scenery, e.g., cities in the Stuttgart area like ``T\"ubingen'',
``Reutlingen'', ``Esslingen'' have the common suffix {\em -ingen}.

The morphological analysis (see section \ref{sec:entag}) operates with
a list of so--called onomastic suffixes to identify place names.

\section{Proper Name Tagging}
\label{sec:entag}

An overview of the tagging process is shown in figure 2.

\begin{figure}[htbf]
\centerline{\psfig{figure=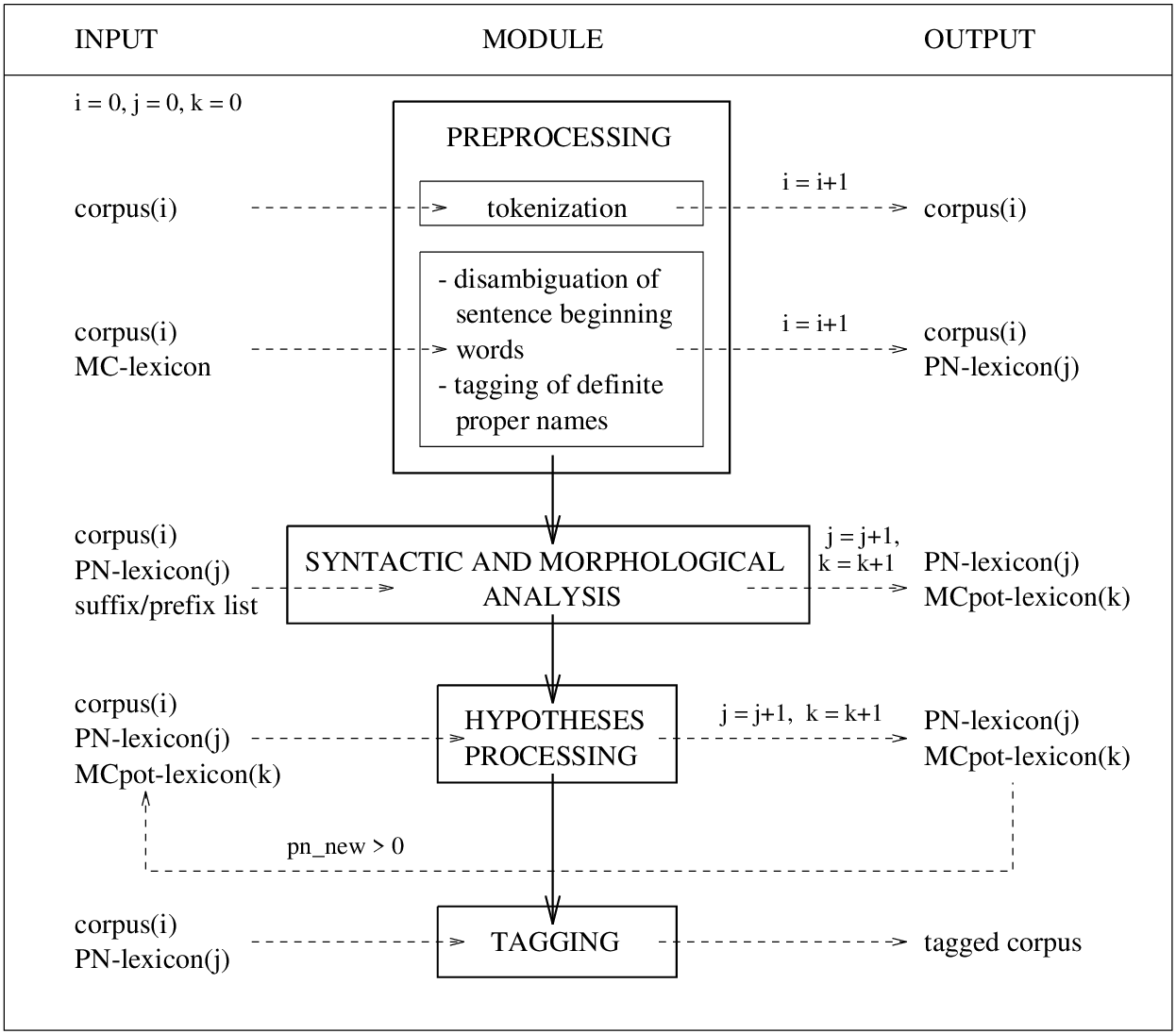,width=7.5cm}}
\captionsenglish{{\small Figure 2: proper name tagging}}
\label{fig:entag}
\end{figure}

\subsection*{Preprocessing}

The corpus has to be preprocessed first of all. This includes the
tokenization of the corpus in which all punctuation marks are
separated from the words to allow the following disambiguation of
sentence--initial words. This disambiguation uses a heuristic derived
from the one used in CLAWS \cite{Garside1987a}: if a sentence--initial
word also occurs inside of a sentence with a lower case initial
letter, then it is not a noun (normal noun or proper name) and represented with
lower case letters. For this I use a list of all words with lower
case initial letter found in the corpus which is stored in an
AVL--tree
\cite{Wirth1983} for better searching and inserting.

After this, a first run through the corpus is done to identify definite
proper names occuring in the contexts of the MC--lexicon. Apart from
appositons as mentioned above, this lexicon contains
speech--embedding (``redeeinbettende'') verbs like ``sagte''and
``fragte'' frequently used in political newspaper texts, as in:

\begin{itemize}
\item[(5)] die Abgeordnete {\em Kelly} sagte, ...\\
	(the member of parliament {\em Kelly} said, ...)
\item[(6)] {\em Heinlein} f\"ugte hinzu, ...\\
	({\em Heinlein} added, ...)
\item[(7)] so fragte {\em Apel}\\
	({\em Apel} asked)
\end{itemize}

The MC-lexicon also contains prepositions and preposition frames to
identify place names, as in:

\begin{itemize}
\item[(8)] bei {\em Frankfurt}\\
	(near {\em Frankfurt})
\item[(9)] aus {\em S\"ollingen} bei {\em Baden--Baden}\\
	(from {\em S\"ollingen} near {\em Baden--Baden})
\item[(10)] im Raum {\em Landshut}\\
	(in the {\em Landshut} area)
\end{itemize}

All proper names are stored in the PN--lexicon which is used during
the entire processing.

\subsection*{Syntactic and Morphological Analysis}
\label{sec:syn}

In the following analysis, the immediate syntactic and morphological
context of all capitalized words is examined. If the capitalized word
is already included in the PN--lexicon, then its immediately preceding
context is stored as a potential minimal context in the MCpot--lexicon
if it comprises one or more capitalized words. Cases where the proper
name is marked as genitive are not considered because this could lead
to wrong MCs (e.g., {\em Aussage W\"orners, Besuch Lafontaines}). The
collection of potential minimal contexts is also done in the
hypotheses processing, which follows. For example, the proper name {\em
W\"orner} supplies the MCs: {\em Bundesverteidigungsminister,
Verteidigungsminister, Minister, Nato--Generalsekret\"ar}.

For the recognition of place names, a suffix list is used containing
onomastic suffixes like {\em --acker, --aich, --beuren, --hafen,
--hausen, --stetten, --weiler} and a prefix list containing prefixes like
{\em Mittel--, Ost--, West--, Zentral--}. In addition to this the ending
of the left capitalized word of two adjacents is checked for
adjectival endings {\em --er, --aner}, as in:

\begin{itemize}
\item[(11)] Mainzer Landtag\\
	(the state parliament of {\em Mainz})
\item[(12)] M\"unsteraner Parteitag\\
	(the party conference of {\em M\"unster})
\end{itemize}

If they also occur without this ending ({\em Mainz, M\"unster}), then
these forms are proper nouns and are stored in the PN--lexicon. The
adjectival forms in (11)--(12) are considered as adjectives (following
\cite{Fleischer1989}, p. 265).

\begin{sloppypar}
Furthermore, loose appositional constructions (``lockere
appositionelle Konstruktionen'',
\cite{Hackel1986}) as in (13)--(14) are analyzed according to the
patterns of noun phrases which occur before the proper name.
\end{sloppypar}

\begin{itemize}
\item[(13)] der Staatssekret\"ar des Landesinnenministeriums, {\em Basten},
...\\
	(the under--secretary of the Department of the Interior, {\em Basten}, ...)
\item[(14)] der Chef des Schweizer Wehrministeriums, Bundesrat {\em Koller},
...\\
	(the director of the Swiss Department of the Armed Forces, the minister of
state {\em Koller}, ...)

\end{itemize}

During this run through the corpus, a second AVL--tree is constructed
in which all capitalized words are stored together with some
information that can be useful for the hypotheses processing. For each
word (node) there is a counter for all occurences of the word with
an article and a list of all its immediately preceding words, if these
are also capitalized or are prepositions (see table 1).

\begin{table}
\begin{tablefont}
\begin{tabular}{|p{1,5cm}|p{4cm}|r|}\hline
{\normalsize Node} & {\normalsize List} & {\normalsize Article}\\\hline\hline
ADN	&	bei	&	0\\
	&	Nachrichtenagentur &	\\\hline
Angaben	&	nach	&	1\\
	&	Donnerstag &	\\\hline
Belgien	&	aus	&	0\\
	&	in	&	\\\hline
Baum	&	FDP-Politker &  0\\
	&	FDP-Abgeordnete & \\\hline
\end{tabular}
\end{tablefont}
\captionsenglish{{\small Table 1: contexts of capitalized words}}
\end{table}

\subsection*{Hypotheses Processing}

In this section of the procedure, hypotheses are generated and
evaluated. A hypothesis may consist of two adjacent capitalized words
or a preposition with a capitalized word. These hypotheses are
evaluated on the basis of all occurences of the second word found in
the corpus.

A hypothesis of two capitalized words is rejected, if

\begin{enumerate}
\item the left word is already in the PN-lexicon
\item the right word is an inflected form which is not possible with PNs.
\end{enumerate}

All other hypotheses are analyzed in the following way. If the left
word is a MCpot or a derived form of a MCpot, then the right word is a
proper name.  For example ``Senatspr\"asident Spadolini'' is analyzed
as proper name ``Spadolini'' with the apposition ``Senatspr\"asident''
which is derived from the MCpot ``Pr\"asident''. The hypothesis is
also accepted if the right word has a genitive ending and occurs
without this ending in the corpus, because only proper names may occur
in such constructions, as in (15). Normal nouns have to be
accompanied by an article, as in (16).

\begin{itemize}
\item[(15)] die Strategie {\em Frankreichs}\\
	(the strategy of {\em France})
\item[(16)] die Strategie des M\"orders\\
	(the strategy of the murderer)
\end{itemize}

A hypothesis of a preposition and a capitalized word is rejected, if
the capitalized word

\begin{enumerate}
\item is a potential minimal context
\item is followed by a genitive article
\item is followed by a past participle.
\end{enumerate}

The latter two conditions exclude such constructions (``feste Syntagmen''),
as in:

\begin{itemize}
\item[(17)] aus Anla{\ss} des\\
	(on the occasion of)
\item[(18)] in Kauf genommen\\
	(accepted)
\end{itemize}

In addition, it is checked whether we have a construction like ``zu
Olims Zeiten'', i.e., whether the capitalized word has a genitive
ending and is followed by a capitalized word. For example, we found
the following proper names:

\begin{itemize}
\item[(19)] in {\em Lafontaines} Worten\\
	(in the words of {\em Lafontaine})
\item[(20)] in {\em Stoltenbergs} Bilanz\\
	(in {\em Stoltenberg's} the balance sheet)
\item[(21)] gegen {\em Hitlers} Erm\"achtigungsgesetz\\
	(against {\em Hitler's} Enabling Act)
\end{itemize}

All resulting hypotheses are evaluated by another procedure which
takes into account the AVL--tree containing all capitalized words
together with the distributional information described above. Because
the corpus is very small and often there is only one occurence of a
word, this information is not very reliable and therefore
error--prone. This could be improved by the application of the
procedure to a very large corpus (several million words). At this
point, it is only checked whether the right word occurs with an
article (a clue for a normal noun) and whether it often occurs
with other capitalized words or prepositions (a clue for a
proper name). Proper names are normally not used with articles with the
exception of ones -- mostly cases place names and
institutional names -- which always occur with an article (e.g. ``die
T\"urkei'', ``die Vereinigten Staaten''). So, this method has to be
used carefully.

The processing of hypotheses is iterated until no more proper names
can be found (pn\_new = 0), because new proper names supply new
contexts and new contexts may supply new proper names.

\subsection*{Tagging}

In order to tag the proper names collected in the EN--lexicon, it is
necessary to run through the corpus for a last time. All words listed
in the EN--lexicon are tagged as proper names.\\

The procedure of proper name tagging was implemented in C under UNIX.

\section{Evaluation}

The first half of the corpus was used to develop the procedure, the
second half served for an evaluation. For the evaluation, all proper
names in the second corpus half were manually tagged and (manually)
compared to the result of the automatic tagging procedure applied to
this corpus part, i.e., to a corpus of 25,000 words.  Of the 1300
proper name tokens 461 occurrences were not recognized, 30 text words
were wrongly tagged as proper names. This corresponds to a recognition
rate of about 65\% (counting errors not excluded). In order to provide
background for this figure, some of the problems are discussed here in
more detail.

The preprocessing module could be improved by enlarging
the MC--lexicon with a list of most frequently used
first names, for example. For the recognition of non--German proper names, it
could be possible to add non--German titles and forms of address as well. The
latter were also found in the corpus (e.g. {\em Captain
Alan Stephenson, Lord Carrington}).

 At the Moment, first names are collected in the MCpot--lexicon if
they are used attributively to a surname already recognized. This is
in contrast to the approaches of
\cite{Fleischer1989} and others (\cite{Wimmer1973},
\cite{Kalverkamper1978}), who analyze first names and surnames as a
unit. One reason for this is that only the surname can be inflected,
as in (22). But as this also applies to titles, as in (23), the reason
does not hold.

\begin{itemize}
\item[(22)] {\em Peter M\"uller\underline{s}} Auto\\
	(the car of {\em Peter M\"uller})
\item[(23)] Minister {\em W\"orner\underline{s}} Rede\\
	(the speech of minister {\em W\"orner})
\end{itemize}

A better argument is that constructions of first name and
surname cannot be expanded, e.g., as loose appositional
constructions.

The procedure of proper name tagging described here is not able to
recognize multi--word proper names because only two adjacent
capitalized words (apposition + proper name) are examined. Table 2
shows an excerpt of unresolved hypotheses in which some multi--word
proper names consisting of first name and surname ({\em Albrecht
M\"uller, Angelika Beer, Harry Ristock, Ruth Winkler, Josef Felder, Gabi
Witt, Florian Gerster, Sepp Binder, Kurt Schumacher}), of
normal nouns ({\em (das) Deutsche Rote Kreuz, Kleine Brogel, Ewige
Lampe}) and of some non-German proper names ({\em Alan Stephenson,
(Canadian) Air Group, Central Enterprise, Frecce Tricolori,
Standardisation Agreement, Acrobatic Full Scale}) are found.

\begin{table}
\begin{tablefont}
\begin{tabular}{|r|p{6,4cm}|}\hline
{\normalsize Text}	& {\normalsize Hypothesis}\\\hline\hline
1 	& Militaerflughafen Rhein-Main\\
2 	& Dutzend Personenwagen\\
2 	& Captain Alan\\
2 	& Alan Stephenson\\
6 	& Mitte April\\
7 	& Metern Abstand\\
11 	& Fraktionskollege Albrecht\\
11 	& Albrecht Mueller\\
12 	& Kanadische Luftwaffendivision\\
12 	& Air Group\\
12 	& Hochleistungsflugzeug F-18\\
13 	& Central Enterprise\\
13 	& Central Enterprise\\
14 	& Central Enterprise\\
22 	& Frecce Tricolori\\
22 	& Deutsche Rote\\
22 	& Rote Kreuz\\
22 	& Dutzend Demonstranten\\
22 	& Autobahnzufahrt Frankfurt-Sued\\
22 	& Luftsportgruppe Breitscheid/Haiger\\
23 	& Kleine Brogel\\
24 	& Fraktionskollegin Angelika\\
24 	& Angelika Beer\\
25 	& Ende September\\
27 	& IG Metall\\
27 	& Harry Ristock\\
27 	& Lehrerin Ruth\\
27 	& Ruth Winkler\\
28 	& Regierung Kohl\\
28 	& Prozent Kandidatinnen\\
30 	& Leitende Oberstaatsanwalt\\
30 	& Oberstaatsanwalt Sattler\\
32 	& Frecce Tricolori\\
34 	& Geburtstag Bert\\
34 	& Josef Felder\\
34 	& Gabi Witt\\
34 	& Ewige Lampe\\
34 	& Museumsdorf Muehlendorf\\
34 	& Florian Gerster\\
34 	& Sepp Binder\\
34 	& Kurt Schumacher\\
35 	& Standardisation Agreement\\
35 	& Standardisation Agreement\\
35 	& Acrobatic Full\\
35 	& Full Scale\\
36 	& Frecce Tricolori\\
36 	& Frecce Tricolori\\
36 	& Frecce Tricolori\\
36 	& Demokratische Proletarier\\
37 	& IG Metall\\
37 	& IG Chemie\\
37 	& IG Bergbau\\
39	& Kanzleramt Erwaegungen\\
56	& Partei Ernst\\
96	& Bundespartei Stellung\\\hline
\end{tabular}
\end{tablefont}
\captionsenglish{{\small Table 2: unresolved hypotheses (excerpt)}}
\end{table}

 The non-German proper names are often put in quotation
marks, so this could be an additional criterion for the hypotheses
evaluation, but cases in which quotation marks are used to emphasize
or to cite one or more words must be excluded (24).

\begin{itemize}
\item[(24)] die FDP warnt vor ``Panikmache''\\
	(the FDP warns of ``panic mongering'')
\end{itemize}

Multi--word proper names consisting of normal nouns or mixed of normal
nouns, adjectives, articles, prepositions and proper names constitute
a major problem. Apart from the fact that adjectives and prepositions
belonging to a proper name are capitalized, some of these proper names
(25) behave like normal nouns, i.e., they are inflectional and take an
article, but some do not (26)-(28). The latter are mostly used with an
introductory apposition and often put in quotation marks. For one it is
difficult to determine which constituents belong to
the proper name, and which do not when the construction can be modified
and reduced as well (e.g. {\em Vereinigte Staaten von Amerika, die
Staaten, die Bundesrepublik, Deutschland}). Under the more
distributional analysis described here, it is not possible to
recognize them and no easy solution is possible. In secondly place,
it is possible to recognize them if we know the minimal context (here
{\em Luftwaffenbasis, Gasthaus, Stra{\ss}e}), which may be
resolved if we use a very large corpus, and if we consider more than
one following word and existing quotation marks.

\begin{itemize}
\item[(25)] {\em die Vereinigten Staaten} und {\em die Bundesrepublik
Deutschland}\\
	({\em the United States} and {\em the Federal Republic of Germany})
\item[(26)] auf der nordbelgischen Luftwaffenbasis {\em Kleine Brogel}\\
	(at the North Belgian air force base {\em Kleine Brogel})
\item[(27)] ein Teil von ihnen geht [...] ins Gasthaus ``{\em Ewige Lampe}''\\
	(some of them go to the inn ``{\em Ewige Lampe}'')
\item[(28)] ich habe in der Stra{\ss}e ``{\em Am Mariahof}'' gewohnt\\
	(I have lived in the street ``{\em Am Mariahof}'')
\end{itemize}

Some of the remaining hypotheses in Table 2 are noun pairs consisting
of quantity terms and normal nouns (29)-(31) or constructions with
month names (32). Quantity terms could be excluded by an exception
list and month names could be added to the EN--lexicon from the start.

\begin{itemize}
\item[(29)] ein Dutzend Personenwagen/Demonstranten\\
	(a dozen automobiles/demonstrators)
\item[(30)] mindestens vierzig Prozent Kandidatinnen\\
	(at least 40 per cent candidates)
\item[(31)] nach Metern Abstand\\
	(after a distance of some metres)
\item[(32)] Mitte April/Ende September\\
	(in the middle of April/at the end of September)
\end{itemize}

 But some of the remaining hypotheses are the result of a free German
word order, often observed in sentences with support verb
constructions (34: {\em Ernst machen mit} (to be serious about), 35:
{\em Stellung beziehen gegen} (to take a stand against)). The hypotheses
`Kanzleramt Erw\"agungen' in sentence (33) could be ruled out if the
form `Erw\"agungen' was analyzed as a non--possible inflection form of a
proper noun and therefore as a normal noun. This was not performed
by the morphological analysis\footnote{%
The analysis is based on a very simple mechanism: inflectional endings
which are not possible for proper names are removed from the word
under consideration, and the remaining form is searched for in the
corpus. If successful, the word cannot be a proper name and the
hypothesis is rejected; if not, the hypothesis is kept.
},
because there were no occurrences of `Erw\"agung' without a plural ending
in the corpus. This could be improved by the use of a very large
corpus or a powerful morphological analyzer (e.g. GERTWOL,
\cite{Koskenniemi1994}). The support verb constructions could be
excluded if we look for typical verbs used in such constructions ({\em
machen, bringen, nehmen, ...}).

\begin{itemize}
\item[(33)] ... war bekanntgeworden, da{\ss} im Kanzleramt Erw\"agungen [...]
stattf\"anden, wie ...\\
	(... became known that the chancellorship takes into consideration ...)
\item[(34)] ... wenn seine Partei Ernst macht mit ...\\
	(... if his party gets serious about ...)
\item[(35)] ... indem man [...] gegen die Bundespartei Stellung bezieht\\
	(... while taking a stand against the federal party)
\end{itemize}

Most of the incorrectly tagged proper names are the result of the
hypotheses processing, because the corpus is too small. For example,
the evaluation of the hypothesis `ohne R\"ucksicht' (with no
consideration) provides `R\"ucksicht' as proper name, because it also
occurs with the preposition `aus' (from), which is frequently used
with place names and never occurs with an article, but its frequency
is only 4. This is not representative for a reliable conclusion and it
is hoped that a very large corpus would allow for a better analysis.

\section{Conclusions and Future Perspectives}

Most of the known statistically based tagging systems are confronted
with the problem of proper name tagging. In German the problem is not
only restricted to the disambiguation of sentence--initial words but
also occurs with sentence--internal capitalized words. The procedure
of proper name tagging described here makes use of a database of
definite minimal contexts as a starting point for an analysis which
takes into account both morphological and syntactic properties of
proper names. Furthermore, this local analysis is supported by a
global analysis regarding all occurrences of capitalized words in the
corpus. This global analysis should be improved by a larger corpus
than the one used, and a more meaningful statistic procedure, like
{\em mutual information} \cite{Church1990b}. However, the central idea
of an incremental procedure for the collection of proper name contexts
is encouraging. It is planned to include this proper name
tagging in the German part-of-speech tagger {\sc Likely}
\cite{Feldweg1993a} developed in T\"ubingen to disambiguate all the
remaining cases where the tagger could not decide between proper name
or normal noun.

\end{document}